\def\year{2020}\relax
\documentclass[letterpaper]{article} % DO NOT CHANGE THIS
\usepackage{aaai20}  % DO NOT CHANGE THIS
\usepackage{times}  % DO NOT CHANGE THIS
\usepackage{helvet} % DO NOT CHANGE THIS
\usepackage{courier}  % DO NOT CHANGE THIS
\usepackage[hyphens]{url}  % DO NOT CHANGE THIS
\usepackage{graphicx} % DO NOT CHANGE THIS
\usepackage{graphicx}  % DO NOT CHANGE THIS
\usepackage{epsfig}
\usepackage{amsmath}
\usepackage{amssymb}
\usepackage{flafter}
\usepackage{bbm}
\usepackage{indentfirst}
\usepackage{color}
\usepackage{tabularx} 
\usepackage{subfigure}
\usepackage{verbatim}
\usepackage{multirow}
\usepackage{makecell}
\usepackage{tabu}
\usepackage[boxed,ruled,commentsnumbered]{algorithm2e}
\newcommand{\RNum}[1]{\uppercase\expandafter{\romannumeral #1\relax}}
\usepackage{enumerate}
\newcommand{\topcaption}{
\setlength{\abovecaptionskip}{-2pt}
\setlength{\belowcaptionskip}{1pt}
\caption}

\title{Discriminative and Robust Online Learning for Siamese Visual Tracking}

\author{\Large \textbf {Jinghao Zhou, Peng Wang, Haoyang Sun} \\
School of Computer Science and School of Automation, Northwestern Polytechnical University, China \\
National Engineering Laboratory for Integrated Aero-Space-Ground-Ocean Big Data Application Technology, China \\
jensen.zhoujh@gmail.com,  peng.wang@nwpu.edu.cn,  sunhaoyang@mail.nwpu.edu.cn}

\begin{document}

\maketitle

%%%%%%%%% ABSTRACT
\begin{abstract}
   The problem of visual object tracking has traditionally been handled by variant tracking paradigms, either learning a model of the object's appearance exclusively online or matching the object with the target in an offline-trained embedding space. Despite the recent success, each method agonizes over its intrinsic constraint. The online-only approaches suffer from a lack of generalization of the model they learn thus are inferior in target regression, while the offline-only approaches (e.g., convontional siamese trackers) lack the target-specific context information thus are not discriminative enough to handle distractors, and robust enough to deformation. 
   Therefore, we propose an online module with an attention mechanism for offline siamese networks to extract target-specific features under $L2$ error. We further propose a filter update strategy adaptive to treacherous background noises for discriminative learning, and a template update strategy to handle large target deformations for robust learning.
   Validity can be validated in the consistent improvement over three siamese baselines: \textit{SiamFC}, \textit{SiamRPN++}, and \textit{SiamMask}. Beyond that, our model based on \textit{SiamRPN++} obtains the best results over six popular tracking benchmarks and can operate beyond real-time.
\end{abstract}

%%%%%%%%% BODY TEXT
\section{Introduction}

\begin{figure}[!t]
\begin{center}
\includegraphics[width=0.95\linewidth]{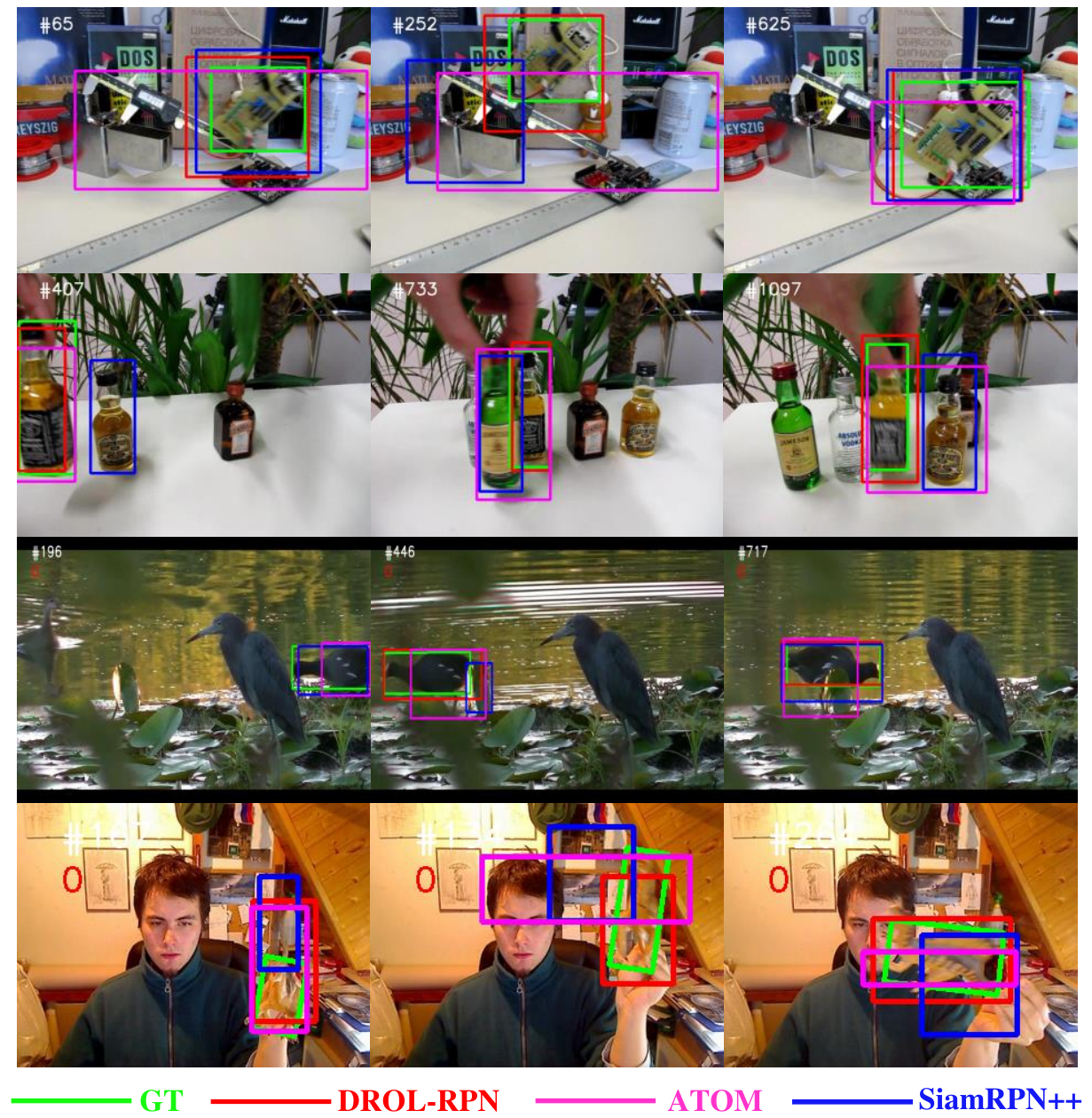}
\end{center}
\caption{Qualitative demonstration and comparison. While accurate bounding box regression eludes \textit{ATOM} and \textit{SiamRPN++} struggles at situations like distractors and deformation, our method \textit{DROL} outperforms \textit{ATOM} with a siamese matching subnet, and {SiamRPN++} with our proposed discriminative and robust online module. }
\label{fig:show}
\end{figure}

Visual object tracking is a fundamental topic in various computer vision tasks, such as vehicle navigation, automatic driving, visual surveillance, and video analytics. Briefly speaking, given the position of an arbitrary target of interest in the first frame of a sequence, visual object tracking aims to estimate its location in all subsequent frames.

In the context of visual object tracking, almost all the state-of-the-art trackers can be categorized into two categories: discriminative and generative trackers. Discriminative trackers train a classifier to distinguish the target from the background, while generative trackers find the object that can best match the target through computing the joint probability density between targets and search candidates. 

Recently, the siamese framework \cite{SiamFC} as a generative tracker has received surging attention in that its performance has taken the lead in various benchmarks while running at real-time speed. Generally, these approaches obtain a similarity embedding space between subsequent frames and directly regress to the real state of the target \cite{SiamRPN,SiamMask}. Despite its recent success, since the siamese paradigm is exclusively trained offline, it suffers from severe limitations: 
\begin{enumerate}[(I)]
\item Siamese approaches ignore the background information in the course of tracking, leading to inferior discriminative capability for target classification in the face of distractors. 
\item Siamese approaches solely utilize the first frame as template or merely update it through averaging the subsequent frames, thus its performance degrades with huge deformation, rotation, and motion blur and results in poor robustness for target regression.
\end{enumerate}
 
Probing into a unified solution to these issues arising from the parameter-fixed inherency of siamese tracking, we're motivated to introduce an online mechanism referring to the discriminative trackers which often feature a discriminative classifier and an efficient online update strategy. The unique strengths in online fashion, that is, the attainment of dynamic video-specific information, have long eluded siamese networks given its offline settings. However, the directly online updating siamese networks may be improper since any introduction of noises during tracking may deteriorate the generative ability of offline-trained embedding space.

Given the above intuitive analysis, a standalone online module that is systematically integrated but mutually independent with siamese networks is proposed as a complementary subnet. The online subnet is designed with an attention mechanism for extraction to most representative target features and is optimized efficiently to cater for time requirements. The response map of the online module is utilized for two earlier-mentioned issues separately. Specifically, limitation (\RNum{1}) is resolved by fusing the online response map with siamese classification score yielding an adaptive classification score for discriminative tracking, while limitation (\RNum{2}) is tackled by giving high-quality frames to siamese networks to update template for robust tracking.

To wrap up, we develop a highly effective visual tracking framework and establish a new state-of-the-art in terms of robustness and accuracy. The main contributions of this work are listed in fourfold:
\begin{itemize}
\item We propose an online module with attention mechanism optimized as a classification problem for siamese visual tracking, which can fully exploit background information to extract target-specific features.
\item We propose discriminative learning using target-specific features via score fusion to help siamese networks handling distractors and background noises. 
\item We propose robust learning using target-specific features via template update to improve their robustness handling deformation, rotation, and illumination, etc.
\item The proposed online module can be integrated with a variety range of siamese trackers without re-training them.  
Our method consistently set new state-of-the-art performance on $6$ popular benchmarks of \textit{OTB100}, \textit{VOT2018}, \textit{VOT2018-LT}, \textit{UAV123}, \textit{TrackingNet}, and \textit{LaSOT}. 
%We propose a parallel framework which can be easily applied to a wide scope of siamese trackers without any re-training. 
\end{itemize}

%------------------------------------------------------------------------
\section{Related Work}

\noindent\textbf{Siamese visual tracking.} Recently, the tracking methods based on siamese network \cite{SiamFC} in the literature greatly outperformed other trackers due to heavy offline training, which largely enriches the depiction of the target. The following works include higher-fidelity object representation \cite{SiamMask,SiamRPN}, deeper backbone for feature extraction \cite{SiamRPN++,SiamDW}, and multi-stage or multi-branch network design \cite{C-RPN,Siam-BM}. Beyond that, another popular optimization is template's update for robust tracking in the face of huge target's appearance change, either by using fused frames \cite{UpdateNet} or separately selected frames \cite{MemTrack}. However, the lack of integrating background information degrades its performance on challenges like distractor, partial occlusions, and huge deformation. In fact, such rigidity is intrinsic given the fact that the object classification score from siamese framework turns out to be messy thus requires a centered Gaussian window to ensure stability. While many works intend to construct a target-specific space, by distractor-aware offline training \cite{DaSiamRPN}, residual attentive learning \cite{RASNet}, gradient-based learning \cite{MLT,TADT}, this embedding space obtained through cross-correlation has yet to explicitly account for the distractors and appearance variation. 

\noindent\textbf{Online learning approach.} Online learning is a dominant feature for discriminative trackers to precede, which is achieved by training an online component to distinguish the target object from the background. Particularly, the correlation-filter-based trackers \cite{KCF,ECO} and classifier-based trackers \cite{TLD,MDNet} are among the most representative and powerful methods. These approaches learn an online model of the object's appearance using hand-crafted features or deep features pre-trained for object classification. Given the recent prevalence of meta-learning framework, \cite{DiMP,Meta-tracker} further learns to learn during tracking. Comparatively speaking, online learning for siamese-network-based trackers has had less attention. While previous approaches are to directly insert a correlation filter as a layer in the forward pass and update it online \cite{CFNet,DSiam}, our work focuses on the siamese's combination with an online classifier in a parallel manner with no need to retrain them.

\noindent\textbf{Target regression and classification.} Visual object tracking synchronously needs target regression and classification, which can be regarded as two different but related subtasks. While object classification has been successfully addressed by the discriminative methods, object regression is recently upgraded by the introduction of Regional Proposal Network \cite{SiamRPN} in siamese network owing to the intrinsic strength for generative trackers to embed rich representation. Nevertheless, the majority of trackers in the gallery stand out usually with the strength within one subtask while degrading within the other. To design a high-performance tracker, lots of works have been focused on addressing these two needs in two separate parts. MBMD \cite{MBMD} uses a regression network to propose candidates and classify them online. ATOM \cite{ATOM} combines a simple yet effective online discriminative classifier and high-efficiency optimization strategy with an IoUNet through overlap maximization, while the regression is achieved through a sparse sampling strategy compared with the dense one via cross-correlation in siamese networks.

\begin{figure*}
\begin{center}
\includegraphics[width=14cm]{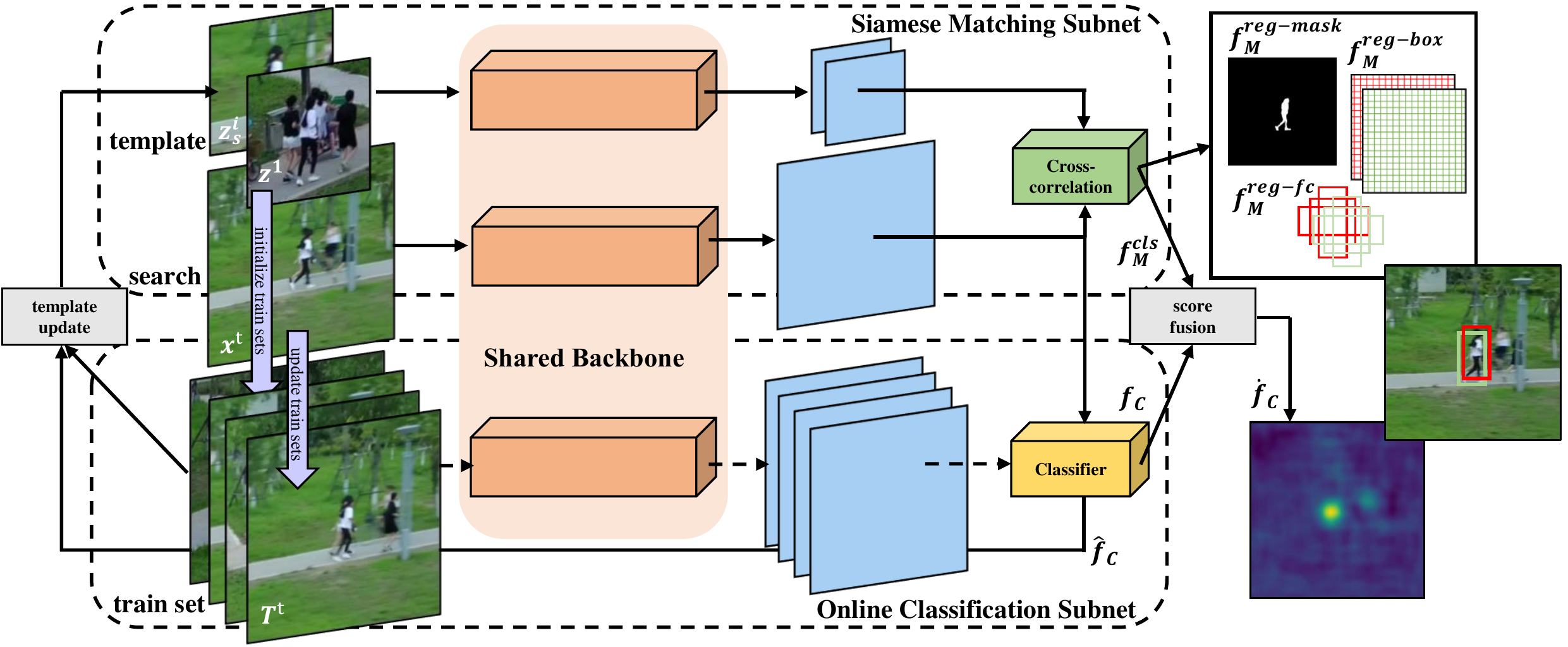}
\end{center}
\caption{Overview of the network architecture for visual object tracking. It consists of siamese matching subnet $M$ majorly accounting for bounding box regression and online classification subnet $C$ generating classification score. Dashed line denotes reference scores passed from the classifier to the updater to select the short-term template $z_s^t$. }
\label{fig:diagram}
\end{figure*}

\section{Proposed Method}

The visual object tracking task can be formulated as a learning problem whose the primary task is to find the optimal target position by minimizing the following objective:
\begin{equation}
L(w)=\sum_{j=1}^m \gamma_j r(f(x_j; w), y_j) + \sum_k \lambda_k \Vert w_k \Vert^2,\label{con:loss}
\end{equation}
where $r(f, w)$ computes the residual at each spatial location and $y_j \in \mathbb{R} ^{W\times H}$ is the annotated label. The impact of each residual $r$ is set by $\gamma_j$ and regularization on $w_k$ is set by $\lambda_k$. 
% In this section, we detail how two subnets in our proposed method as shown in Figure \ref{fig:diagram} can be unified to this formula and complement each other to form a high-performance tracker.

\subsection{Siamese Matching Subnet} 

Before moving to our approach, we firstly review the siamese network baseline. The siamese-network-based tracking algorithms take (\ref{con:loss}) as a template-matching task by formulating as a cross-correlation problem $f_R$ and learn a embedding space $\phi(\cdot)$ that computes where in the search region can best match the template, shown as 
\begin{equation}
f_{M}^{cls}(x,z) = \phi(x)*\phi(z) + b* \mathbbm{1},\label{con:rmodel}
\end{equation}
where one branch learns the feature representation of the target $z$, and the other learns the search area $x$. 

While the brute-force scale search to obtain target regression $f_{M}^{reg}(x,z;w)$ is insufficient in accuracy, siamese network extensions explicitly complement themselves with a Region Proposal Network (RPN) head or a mask head by separately regressing the bounding box and the mask of the target by encoding a subspace for box $[\cdot]^{reg-box}$ and mask $[\cdot]^{reg-mask}$. They each output the offsets for all prior anchors or a pixel-wise binary mask. These 3 variants of target regression can be formulated as:
\begin{equation} 
\begin{split}
\footnotesize
f_{M}^{reg-fc}(x,z) &= \phi(x)*\phi(\hat{z}), \hat{z} \in \{z^i|i=1,2,3\} \\
f_{M}^{reg-box}(x,z) &= [\phi(x)]^{reg-box}*[\phi(z)]^{reg-box}\\
f_{M}^{reg-mask}(x,z) &= [\phi(x)]^{reg-mask}*[\phi(z)]^{reg-mask}
\end{split}
\label{con:or}
\end{equation}
which is seperately adopted by our chosen baseline \textit{SiamFC}, \textit{SiamRPN++}, and \textit{SiamMask}. $\hat{z}$ is the scaled template with maximal response value and $f_{M}^{reg-fc}(x,\hat{z})$ is a $1d$ vector, directly giving target center at pixel with maximal score. $f_{M}^{reg-box}(x,z)$ is a $4d$ vector which stands for the offsets of center point location and scale of the anchors to groudtruth. $f_{M}^{reg-mask}(x,z)$ is a $(63\times63)d$ vector encoding the spatial distribution of object mask.

The eventual predicted target state $s$ is obtained using the same strategy as in \cite{SiamFC,SiamRPN,SiamRPN++,SiamMask}. Additionally, the siamese network takes the residuals for classification commonly as cross-entropy loss, bounding box regression as smooth $L_1$ loss, and mask regression as logistic regression loss. 

\subsection{Target-specific Features} \label{sec:oc}
\begin{figure}[t]
\begin{center}
\includegraphics[width=1.0\linewidth]{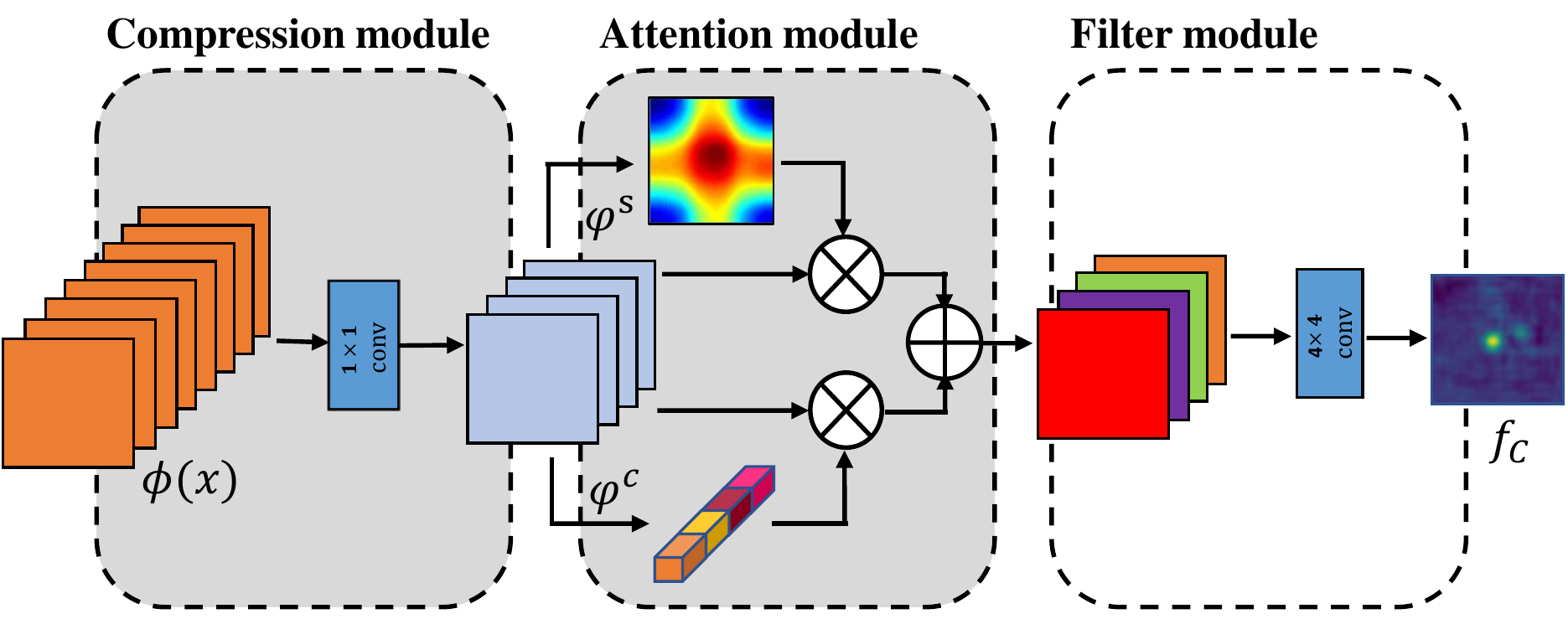}
\end{center}
\caption{Network architecture of proposed online module, which consists of compression module to refine the feature map from backbone suitable for classification, attention module resolving data imbalance and filter module. $\varphi^s$ denotes spatial attention consisting of $2$ FC layers after global average pool (GAP) and $\varphi^c$ channel-wise attention composed of a softmax after channel averaging.}
\label{fig:onlineclassification}
\end{figure}

%\IncMargin{1em}
\begin{algorithm}[t]
%\SetAlgoNoLine 
\SetKwInOut{Input}{\textbf{Input}}
\SetKwInOut{Output}{\textbf{Output}} 
\Input{Video frames $f^1,...,f^t$ of length $L$\;\\ Initial target state $s^1$;}
\Output{Target state $s^t$ in subsequent frames; }
\BlankLine
%\textit{// Initialize}\\
Setting train set $T$ for online classification subnet $C$ and update with augmented search $\{\widetilde{x}^1\}$ \;
Setting template $z, z_s$ for siamese matching subnet $M$\;

\For {t $=$ 2 to $L$}
{Obtain the search $x^t$ based on the previous predicted target state $s^{t-1}$\;

\BlankLine
\textit{// Track}\\
Obtain $f_{C}(x^t) = C(x^t)$ via classification subnet\;
Obtain $f_{M}^{cls}(x^t), f_{M}^{reg}(x^t) = R(x^t)$ using (\ref{con:rmodel})\;
Obtain current target state $s^t$  based on $\dot{f}_{C}(x^t)$ using (\ref{con:omodel})\ and $f_{M}^{reg}(x^t)$;

\BlankLine
%\textit{// Update $C$}\\
\If {${\delta}_c(s^t) \geq \epsilon $}
{$T^{t} \leftarrow T^{t-1} \cup x^t $\;
Train $C$ with $T^t$\;}

\BlankLine
%\textit{// Update $M$}\\
\If {$(t$ mod $T) = 0$}
{$z_{s} \leftarrow z^{t}_s $ using (\ref{con:update})\;
}
}
\caption{Tracking algorithm \label{al}}
\end{algorithm}
%\DecMargin{1em}

The features provided by siamese matching subnet which are trained offline and set fixed to ensure tracking stability and efficiency. However, they're not target-specific and adaptive to the target's domain. Inspired by discriminative trackers, in the initialization stage (the first frame), we propose to harness target-specific features on top of siamese features in a supervised manner by setting (\ref{con:loss}) as $L2$ loss yielding, 

\begin{equation}
r_{C}(f, y_1) = \Vert f_{C} - y_1\Vert^2,\label{con:loss1}
\end{equation}
where the expected output $f_{C}$  is the classification scores for each location belonging to the region of target and overall representing a spatial distribution of labels. $y_1$ is set to Gaussian according to the given target bounding box.

 In light of related researches \cite{TADT,HASiam}, the fitting of confidence scores for all negative background pixels dominates online learning by directly using the standard convolutions, as only a few convolutional filters play an important role in constructing each feature pattern or object category.  The aforementioned data imbalance in both spatial and channel-wise fashion degrades the model's discriminative capabilities. 
 
 To accommodate this issue, we introduce a dual attention mechanism to fully extract target-specific features. As shown in Figure \ref{fig:onlineclassification}, the compression module, and the attention module together form a  target-specific feature extractor. Note that these $2$ modules (gray area) are only fine-tuned with the first frame of given sequences and are kept fixed during tracking to ensure stability. The harnessed target-specific features are then leveraged to optimize the filter module (white area) in subsequent frames.

\subsection{Discriminative Learning via Filter Update}

Extensive study \cite{DaSiamRPN} has proved siamese trackers can be easily distracted by a similar object during tracking even with distractor-aware (target-specific) features. A more profound reason for such inferiority is that there's no online weight update performed to suppress the treacherous background noises.

As a solution, by changing $y_1$ in (\ref{con:loss1}) to $y_i$ centered at predicted location, the update of filter module can be iteratively optimized as tracking proceeds. To further accelerate optimization, reformulating (\ref{con:loss}) into the squared norm of the residual vector $L(w) = \Vert r(w) \Vert^2$, where $r_j(w) = \sqrt[]{\gamma_j}(f(x_j)-y_j)$ and $r_{m+k}(w) = \sqrt[]{\lambda_k}w_k$, induces a positive definite quadratic problem. Instead of using the standard stochastic gradient descent (SGD) as the optimization strategy, following \cite{ATOM}, we used Conjugate Gradient Descent better suited for solving quadratic problem in terms of convergence speed. This algorithm allows an adaptive search direction $p$ and learning rate $\alpha$ during backpropogation in an iterative form.  

The online classification score from the filter module is resized using cubic interpolation to the same spatial size as the  siamese classification score and then fused with it through weighted sum yielding an adaptive classification score, which can be formulated as:
\begin{equation}
\dot{f}_{C}(x;w) = \lambda f_{C}(x;w) + (1-\lambda)f_{M}^{cls}(x,z;w),\label{con:omodel}
\end{equation}
where $\lambda$ is the impact of online confidence score. 

\subsection{Robust Learning via Template Update}

\begin{figure*}[htbp]
\begin{center}
\includegraphics[width=17cm]{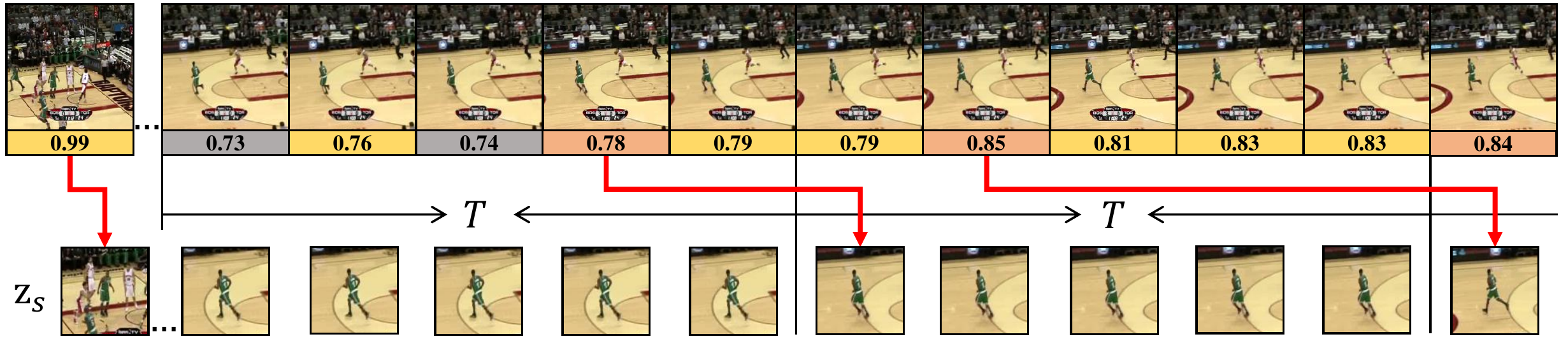}
\end{center}
\caption{Visualized process of proposed template update strategy on sequence \textit{Basketball} of OTB2015. The frame with the highest score in the historical frames will be selected to update the short-term template $z_s$.}
\label{fig:siameseregression}
\end{figure*}

A crucial section of siamese trackers is the selection of templates, as the variation of target appearances often renders poor target regression and even target drift. We thus design an extra template branch for siamese networks to reserve the target information on the most recent frame. The template for this branch is denoted as $z_s$, representing the template in short-term memory as opposed to retaining the first frame for long-term memory. 

It's worth noting that the key for good performance rests largely on the quality of identified templates, thus the selection of possible template candidates need to be designed reasonably. We use the maximal score $\hat f_{C}(x;w)$ of the output from online classifier as a yardstick to select high-quality templates, which can be formulated by
\begin{equation}
z_s = \mathrm{warp}(\arg \mathop{\max}_{x}\hat f_{C}(x;w)* \mathbbm{1}_{\hat f_{C}>\tau_c}),
\label{con:update}
\end{equation}
where $\mathrm{warp}(\cdot)$ is the cropping operation to feature size of the template and $\tau_c$ is filtering threshold suppressing low-quality template facing occlusion, motion blur, and distractors. 

By shifting to the short-term template, $z$ in (\ref{con:rmodel}) can be modified to $z^\prime$ which is decided with
\begin{equation}
z^\prime=\left\{
\begin{array}{rcl}
z_s, & & {\mathrm{IoU}(\hat f_M^{reg}(z_s), \hat f_M^{reg}(z^1)) \geq \upsilon_r,} \\
& &  {\hat f_M^{cls}(z_s) - \hat f_M^{cls}(z^1) \geq \upsilon_c}, \\
z^1, & & \mbox{otherwise},
\end{array} \right.
\end{equation}
where $\hat f_M^{cls}$ denotes location with maximal score and $\hat f_M^{reg}$ its corresponding bounding box. This is to avoid bounding box drift and inconsistency using the short-term template. The full pipeline of our method is detailed in Algorithm \ref{al}.

\section{Experiments}

\subsection{Implementation Details}

Our method is implemented in Python with PyTorch, and the complete code and video demo will be made available at \url{https://github.com/shallowtoil/DROL}.

% Our method is implemented in Python with PyTorch, and the complete code and video demo will be made available at \href{https://github.com/shallowtoil/DROL}{https://github.com/shallowtoil/DROL}. 

\noindent\textbf{Architecture. } Corresponding to the regression variants as described in (\ref{con:or}), we apply our method in three siamese baselines \textit{SiamFC}, \textit{SiamRPN++}, and \textit{SiamMask} yielding \textit{DROL-FC}, \textit{DROL-RPN}, \textit{DROL-Mask} respectively. In \textit{DROL-FC}, we use the modified AlexNet as the backbone and up-channel correlation. In \textit{DROL-RPN}, we use the modified AlexNet and layer $2,3,4$ of ResNet50 and depth-wise correlation. While in \textit{DROL-Mask}, we use layer $2$ of ResNet50 and depth-wise correlation. For the classification subnet, the first layer is a $1\times1$ convolutional layer with ReLU activation, which reduces the feature dimensionality to $64$. The last layer employs a $4\times4$ kernel with a single output channel.

\noindent\textbf{Training phase. } For offline training, since we directly use the off-the-shelf siamese models, no extra offline training stage for the whole framework is required. The training data for our models thus is completely the same as those variants' baselines. For online tuning, we use the region of size $255 \times 255$ of the first frame to pre-train the whole classifier. Similar to \cite{ATOM}, we also perform data augmentation to the first frame of translation, rotation, and blurring, yielding total $30$ initial training samples.

\noindent\textbf{Discriminative learning. }  We add the coming frames to initial training samples with a maximal batch size of $250$ by replacing the oldest one. Each sample is labeled with a Gaussian function centered at predicted target location. We discard the frames with the occurrence of distractors or target absence for filter update. The classifier is updated every $10$ frame with a learning rate set to $0.01$ and doubled once neighboured distractors are detected. To fuse classification scores, we set $\lambda$ to $0.6$ in \textit{DROL-FC} and $0.8$ in \textit{DROL-RPN} and \textit{DROL-Mask}. 

\noindent\textbf{Robust learning. } For template update, to strike a balance between stability and dynamicity when tracking proceeds, we update the short-term template every $T=5$ frames, while $\tau_c$, $\upsilon_r$, and $\upsilon_c$ are set to $0.75$, $0.6$, and $0.5$ respectively. The above hyper-parameters are set using VOT2018 as the validation set and are further evaluated in Section \ref{ablation}.

\begin{table}[!t]
\newcommand{\tabincell}[2]{\begin{tabular}{@{}#1@{}}#2\end{tabular}}
\setlength{\tabcolsep}{0.8mm}{
\begin{center}
\begin{tabular}{|c|c|c|c|c|c|c|}
\hline
\multirow{2}{*}{Method} & \multirow{2}{*}{Backbone} & \multicolumn{2}{c|}{OTB2015} & \multicolumn{2}{c|}{VOT2018} & \multirow{2}{*}{FPS}  \\
\cline{3-6}
& & AUC & Pr & EAO & A  & \\
\hline\hline
% CCOT & VGGNet& 0.682 & 0.903 & 0.267 & 0.494 & 0.3  \\
ECO & VGGNet & 0.694 & 0.910 & 0.280 & 0.484 & 8 \\
UPDT & ResNet50 &{\color{blue} 0.702} & - & 0.378 & 0.536 & - \\
% LADCF & VGGNet & 0.696 & 0.906 & 0.389 & 0.503 & 10 \\
%\hline\hline
MDNet & VGGNet & 0.678 & 0.909 & - & - & 1  \\
ATOM & ResNet18 & 0.669 & 0.882 & 0.401 & 0.590 & 30  \\
DiMP & ResNet50 & 0.684 & 0.894 & {\color{blue} 0.440} & 0.597 & 40  \\
\hline\hline
SiamFC & AlexNet &0.582 & 0.771 & 0.206 & 0.527 & 120  \\
\bf{DROL-FC}& AlexNet & 0.619 & 0.848 & 0.256 & 0.502 & 60 \\
\hline\hline
SiamRPN & AlexNet &0.637 & 0.851 &  0.326 & 0.569 & {\color{red}200}  \\
DaSiamRPN & AlexNet & 0.658 & 0.875 & 0.383 & 0.586 & {\color{blue}160} \\
% C-RPN & AlexNet & 0.663 & 0.882 & - & - & 32  \\
% SPM & AlexNet & 0.687 &0.899  & - & - & 120  \\
\multirow{2}{*}{SiamRPN++} & AlexNet & 0.666 & 0.876 & 0.352 & 0.576 & 180  \\
 & ResNet50 &0.696 & {\color{blue} 0.915} & 0.414 & 0.600 & 35  \\

\multirow{2}{*}{\bf{DROL-RPN}} & AlexNet & 0.689 & 0.914 & 0.376 & 0.583 & 80 \\
& ResNet50 & {\color{red} 0.715} & {\color{red} 0.937} & {\color{red}0.481} & {\color{red}0.616} & 30 \\
\hline\hline
SiamMask & ResNet50 & - & - & 0.347 & 0.602 & 56 \\
\bf{DROL-Mask} & ResNet50 & - & -  & 0.434 & {\color{blue} 0.614} & 40 \\
\hline
\end{tabular}
\end{center}}
\topcaption{state-of-the-art comparison on two popular tracking benchmarks OTB2015 and VOT2018 with their running speed. AUC: area under curve; Pr: precisoin; EAO: expected average overlap; A: accuracy; FPS: frame per second. The performance is evaluated using the best results over all the settings proposed in their original papers. The speed is tested on Nvidia GTX 1080Ti GPU. }
\label{benchmark}
\end{table}

\subsection{Comparison with state-of-the-art}

 We first showcase the consistence improvement of DROL with each of its baseline on two popular benchmarks OTB2015 \cite{OTB2015} and VOT2018 \cite{VOT2018}, as shown in Table \ref{benchmark}. We further verify our best model \textit{DROL-RPN} with backbone ResNet50 on other four benchmarks: VOT2018-LT, UAV123 \cite{UAV123}, TrackingNet \cite{TrackingNet}, and LaSOT \cite{LaSOT}. 

\begin{figure}[!t]
%\centering
\setlength{\abovecaptionskip}{0.cm}
\setlength{\belowcaptionskip}{-0.cm}
\subfigure{
\begin{minipage}[t]{0.22\textwidth}
%\centering
\includegraphics[width=1.6in]{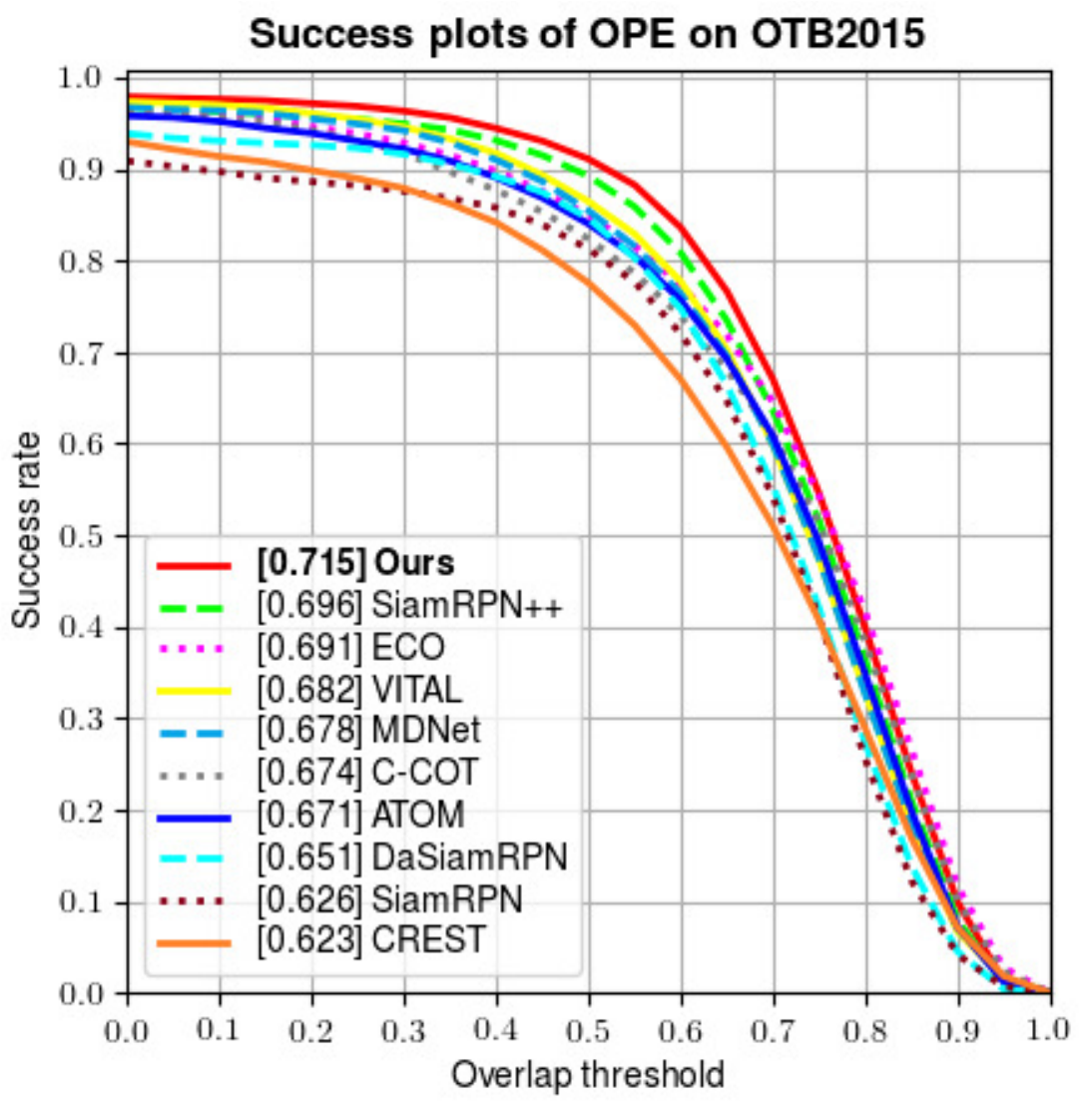}
\end{minipage}
}\subfigure{
\begin{minipage}[t]{0.22\textwidth}
%\centering
\includegraphics[width=1.6in]{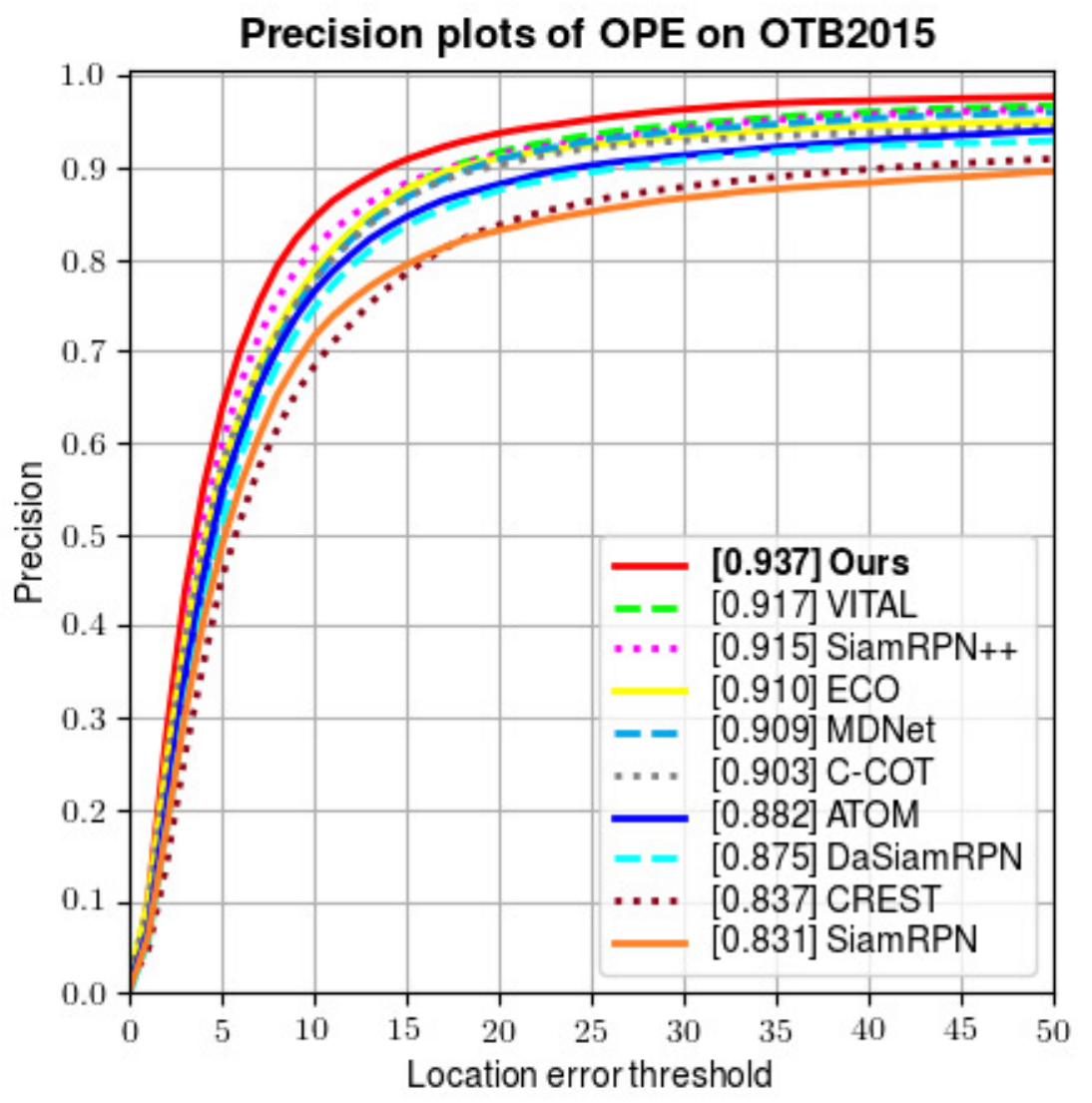}
\end{minipage}
}
%\centering
\caption{state-of-the-art comparison on the OTB2015 dataset in terms of success rate (left) and precision (right).} 
\label{OTB2015}
\end{figure}

\noindent\textbf{OTB2015. } We validate our proposed tracker on the OTB2015 dataset, which consists of 100 videos. Though the latest work SiamRPN++ unveiling the power of deep neural network, as shown in Figure \ref{fig:show},  the best siamese trackers still suffer the agonizing pain from distractors, full occlusion and deformation in sequences like \textit{Baseketball}, \textit{Liquor} and \textit{Skating1}. Comparatively speaking, our approach can robustly discriminate the aforementioned challenges thus obtain a desirable gain above the siamese baseline through online learning. Figure \ref{OTB2015} and Table \ref{benchmark} show that DROL-RPN achieves the top-ranked result in both AUC and Pr. Compared with previous best tracker UPDT, we improve $1.3\%$ in overlap. While compared with ATOM and SiamRPN++, our tracker achieves a performance gain of $4.6\%$, $1.9\%$ in overlap and $5.5\%$, $2.2\%$ in precision respectively. 

\begin{figure}[!t]
\setlength{\abovecaptionskip}{0.cm}
\setlength{\belowcaptionskip}{-0.cm}
\begin{center}
\subfigure[Baseline]{
\begin{minipage}[t]{0.22\textwidth}
%\centering
\includegraphics[width=1.0\linewidth]{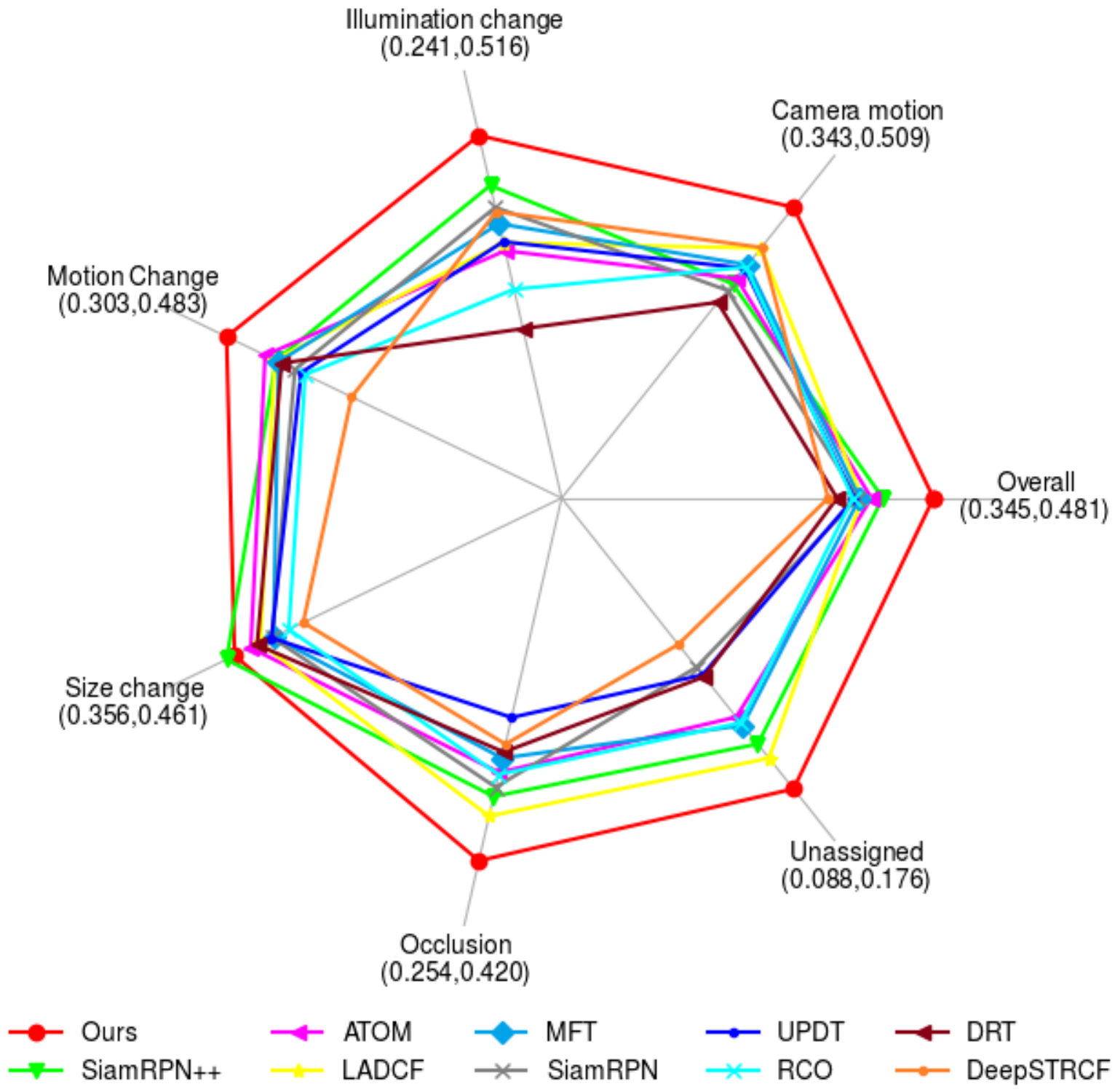}
\end{minipage}
\label{VOT2018b}
}\subfigure[Long-term]{
\begin{minipage}[t]{0.22\textwidth}
%\centering
\includegraphics[width=1.0\linewidth]{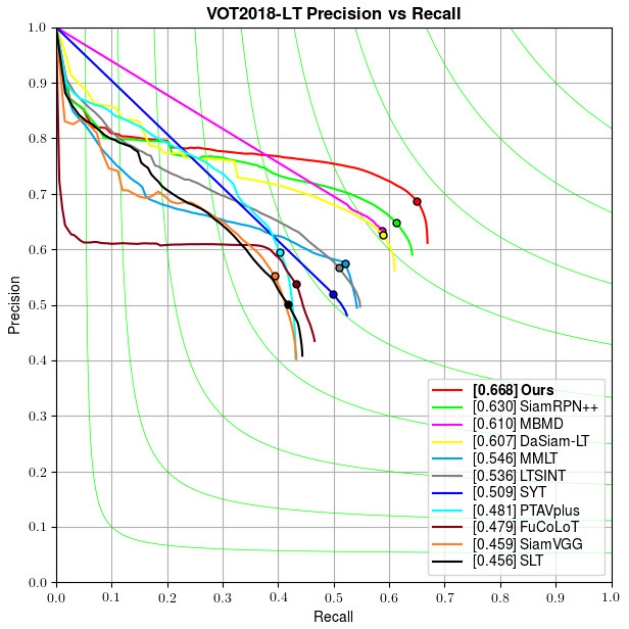}
\end{minipage}
\label{VOT2018l}
}
\end{center}
\caption{state-of-the-art comparison on VOT2018 Baseline (left) and Long-term (right) dataset. In right one, the maximal and minimal value for each attribute is shown in pair, and the rest are permuted propotionally. In left one, the average precision-recall curves are shown above, each with its maximal F-score pointed out at corresponding point.}
\label{fig:VOT2018EAO}
\end{figure}

\begin{table}[!t]
\footnotesize
\setlength{\tabcolsep}{0.8mm}{
\begin{center}
\begin{tabular}{lccccc}  
\hline
  & MMLT & DaSiam-LT & MBMD & SiamRPN++ & \bf{DROL-RPN} \\  
\hline  
%F$\uparrow$  & 0.546 & 0.610 & 0.611 & {\color{blue} 0.630} & {\color{red} 0.668} \\  
Pr$\uparrow$  & 0.574 & 0.627 & 0.642 & {\color{blue} 0.649} & {\color{red} 0.687} \\ 
Re$\uparrow$  & 0.521 & 0.588 & 0.583 &{\color{blue} 0.612} & {\color{red} 0.650} \\   
\hline  
\end{tabular}
\end{center}}
\topcaption{state-of-the-art comparison on the VOT2018-LT dataset in terms precision (Pr) and Recall (Re).} 
\label{VOT2018LT}
\end{table} 

\noindent\textbf{VOT2018. } Compared with OTB2015, VOT2018 benchmark includes more challenging factors, thus may be deemed as a more holistic testbed on both accuracy and robustness. We validate our proposed tracker on the VOT2018 dataset, which consists of $60$ sequences.

As Figure \ref{VOT2018b} shown, we can observe that our tracker DROL-RPN improves the ATOM and SiamRPN++ method by an absolute gain of $8.0\%$ and $6.6\%$ respectively in terms of EAO. Our approach outperformed the leading tracker DiMP by $4.8\%$ in overlap and $1.9\%$ in accuracy. 
% Additionally, DROL-RPM achieves a robustness of $0.122$ and reach a substantial gain of $1.8\%$ compared with VOT2018 challenge winner MFT in terms of robustness after adopting the template update strategy. 

\noindent\textbf{VOT2018-LT. } VOT2018 Long-term dataset provides a challenging yardstick on robustness as the target may leave the field of view or encounters full occlusions for a long period. 
% We validate our method on VOT2018-LT, composed of $35$ videos with an average sequence length of about $2000$ frames. The performance measures precision, recall, and F-score during tracking, 
We report these metrics compared with the state-of-the-art method in Figure \ref{VOT2018l} and Table \ref{VOT2018LT}.  We directly adopting the long-term strategy the same as SiamRPN++ \cite{SiamRPN++} in our siamese subnet. When the target absence is detected and the search region is enlarged, we don't add any search at these moments into training sets to train the classifier. Compared with our baseline SiamRPN++, our approach achieves a maximum performance gain of $3.8\%$ in precision, $3.8\%$ in recall, and $3.8\%$ in F-score respectively. Results show that even encountered huge deformation and long-term target absence, the online classification subnet can still perform desirably. 

\begin{table}[!t]
\footnotesize
\setlength{\tabcolsep}{0.4mm}{
\begin{center}
\begin{tabular}{lcccccc}  
\hline
& ECO & SiamRPN & SiamRPN++ & ATOM & DiMP & \bf{DROL-RPN} \\  
\hline  
AUC  & 0.525 & 0.527 & 0.613 & 0.631 & {\color{blue} 0.643} & {\color{red} 0.652} \\  
Pr & 0.741 & 0.748 & 0.807 & 0.843 & {\color{blue} 0.851} & {\color{red} 0.855} \\ 
\hline  
\end{tabular}
\end{center}}
\topcaption{state-of-the-art comparison on the UAV123 dataset in terms of success (AUC), and precision (Pr).}  
\label{UAV123}
\end{table} 

\noindent\textbf{UAV123.} We evaluate our proposed tracker on the UAV123 dataset. The dataset contains more than $110K$ frames and a total of $123$ video sequences, each with on average $915$ frames captured from low-altitude UAVs.  Table \ref{UAV123} showcases the overlap and precision of the compared tracker. We compare our trackers with previous state-of-the-arts and results show that our approach outperforms the others by achieving an AUC score of $65.7\%$. Our tracker outperforms the SiamRPN++ baseline by a large margin of $4.4\%$ in AUC score and $4.5\%$ in precision. 

\begin{table}[!t]
\footnotesize
\setlength{\tabcolsep}{0.7mm}{
\begin{center}
\begin{tabular}{lcccccc}  
\hline
  & MDNet  & ATOM & SPM & SiamRPN++ & DiMP & \bf{DROL-RPN} \\  
\hline  
AUC & 0.614  & 0.703 & 0.712 & 0.733 & {\color{blue} 0.740} & {\color{red} 0.746} \\  
Pr & 0.555  & 0.648 & 0.661 & {\color{blue} 0.694} & 0.687 & {\color{red} 0.708} \\ 
NPr & 0.710 & 0.771 & 0.778 & 0.800 & {\color{blue} 0.801} & {\color{red} 0.817} \\   
\hline  
\end{tabular}
\end{center}}
\topcaption{state-of-the-art comparison on the TrackingNet \textit{test} dataset in terms of success (AUC), precision (Pr) and normalized precision (NPr). }  
\label{TrackingNet}
\end{table} 

\begin{figure}[!t]
%\centering
\setlength{\abovecaptionskip}{0.cm}
\setlength{\belowcaptionskip}{-0.cm}
\subfigure{
\begin{minipage}[t]{0.22\textwidth}
%\centering
\includegraphics[width=1.6in]{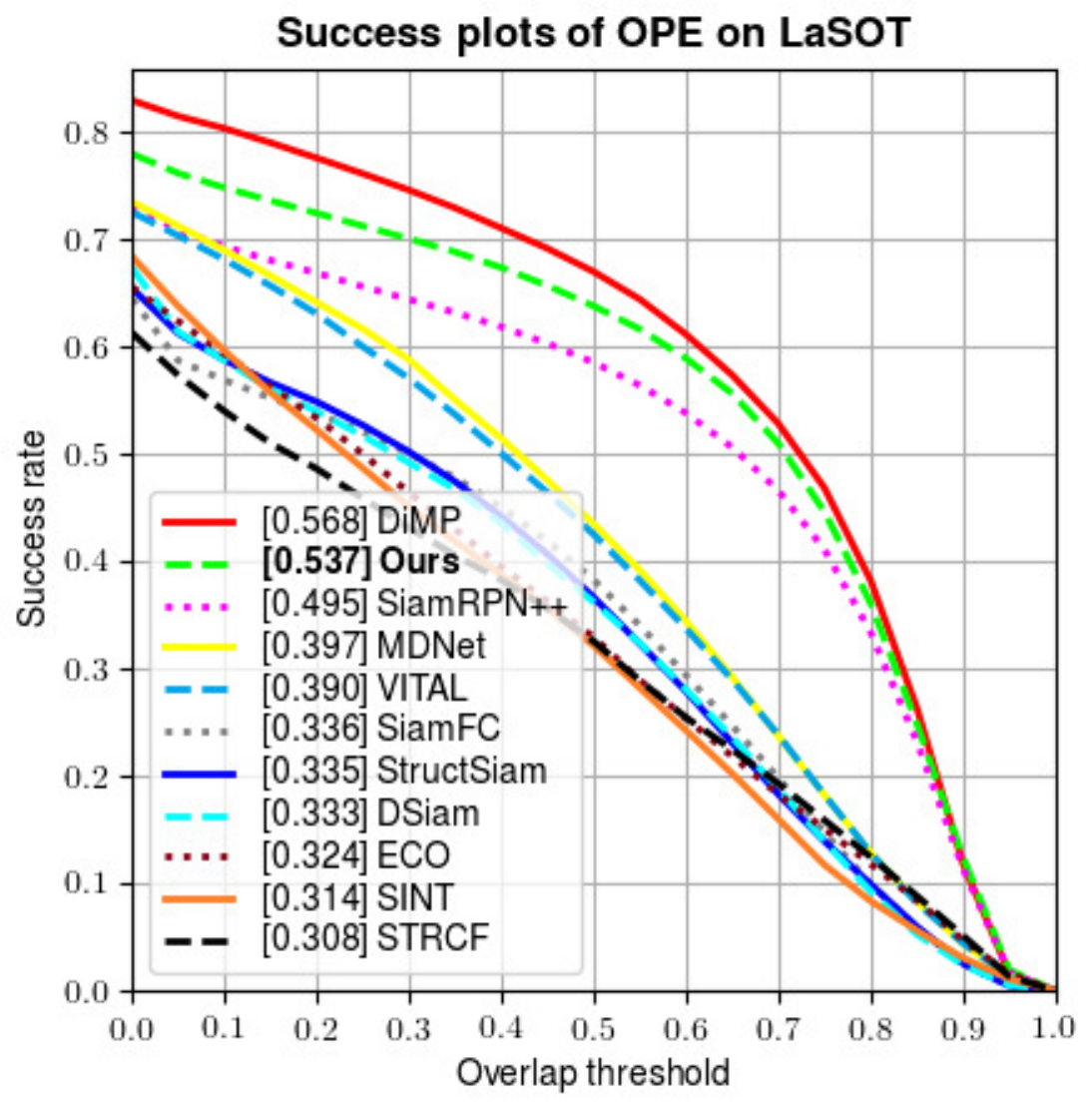}
\end{minipage}
}\subfigure{
\begin{minipage}[t]{0.22\textwidth}
%\centering
\includegraphics[width=1.6in]{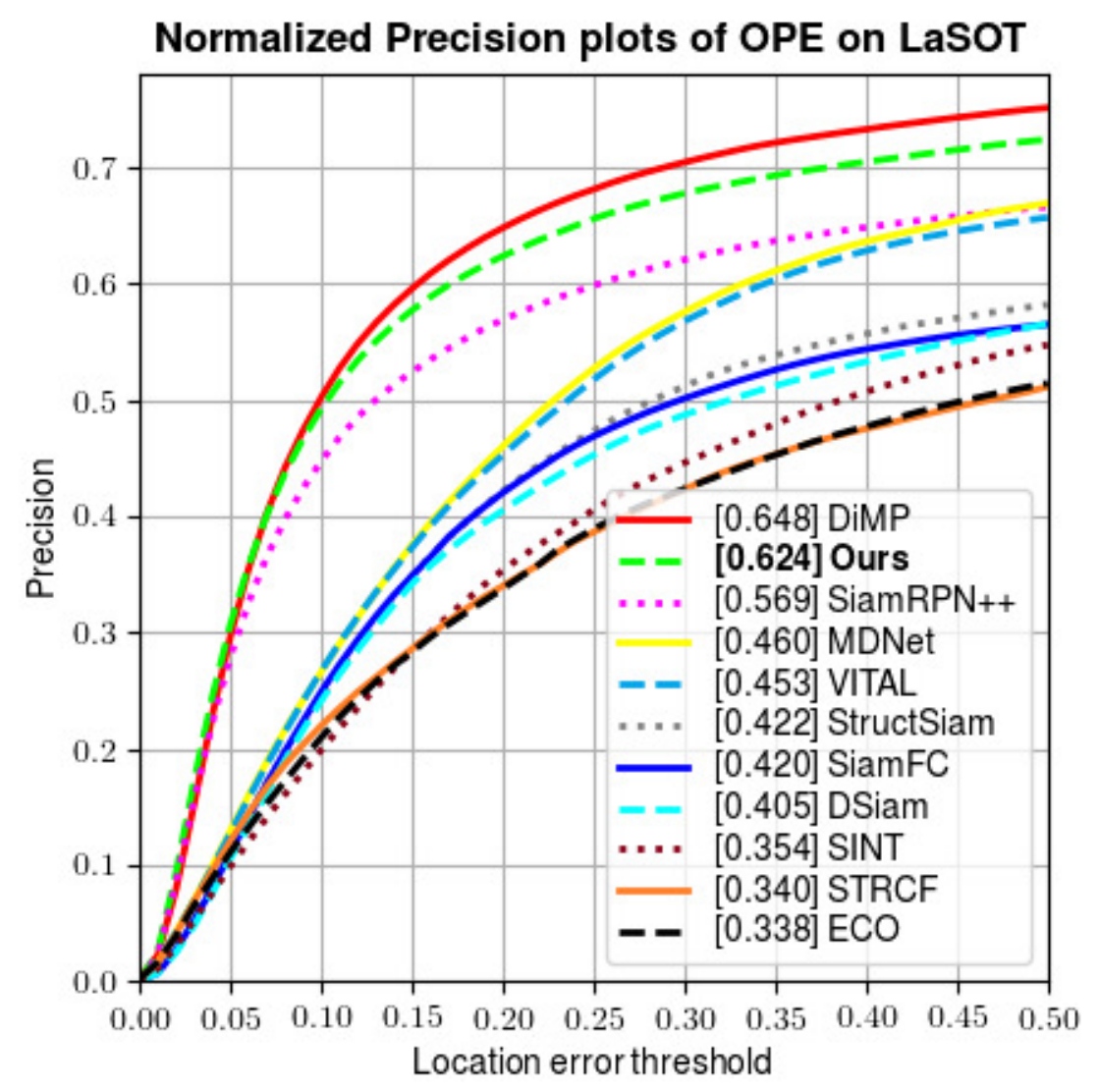}
\end{minipage}
}
%\centering
\caption{state-of-the-art comparison on LaSOT dataset in terms of success rate (left) and normalized precision (right).} 
\label{LaSOT}
\end{figure}

\noindent\textbf{TrackingNet.} We validate our proposed tracker on the TrackingNet \textit{test} dataset, which provides more than $30K$ videos with $14$ million dense bounding box annotations.
% The dataset covers a wide selection of object classes in a broad and diverse context. Its \textit{test} subset composed of $511$ videos allows us access to fully comparing all the trackers in large-volume provided data including common objects in the wild. 
As illustrated in Table \ref{TrackingNet}, our approach DROL-RPN outperforms all previous methods and improves the performance of state-of-the-art by $0.6\%$ in overlap, $1.4\%$ in precision, and $1.6\%$ in normalized precision. 

\noindent\textbf{LaSOT.} We validate our proposed tracker on the LaSOT dataset to further validate its discriminative and generative capability. 
% The large-scale $280$ sequences in \textit{test} set to provide a comprehensive yardstick for evaluating the trackers' overall performances. LaSOT benchmark has an average sequence length of $2500$ frames, which indicates that the model's discriminative and generative capability to deal with the distractor and huge deformation is crucially significant. 
Figure \ref{LaSOT} reports the overall performance of DROL-RPN on LaSOT testing set. Specifically, our tracker achieves an AUC of $53.7\%$ and an NPr of $62.4\%$, which outperformed ATOM and SiamRPN++ without bells and whistles. 

\subsection{Ablation Study}

In this part, we perform ablation analysis to demonstrate the impact of each component in groups and illustrate the superiority of our method (only \textit{DROL-RPN} with backbone ResNet50 is showcased for brevity). Since we search the hyper-parameters using VOT2018 (validation set), we report the performance with LaSOT and TrackingNet (test set).

\begin{table}[!t]
%\scriptsize
\setlength{\tabcolsep}{1.3mm}{
\begin{center}
\begin{tabu}{c|c|c|c|c|c|c|c} 
\multirow{2}{*}{group} & \multirow{2}{*}{$\lambda$} & \multirow{2}{*}{pipeline}  & \multirow{2}{*}{$T$} & \multicolumn{2}{c|}{LaSOT} & \multicolumn{2}{c}{TrackingNet} \\  
\cline{5-8}
&&&& AUC & NPr & AUC & NPr \\
\Xhline{1.2pt} 
\multirow{7}{*}{G1} 
& $0.0$ & \multirow{7}{*}{\textit{Cmp}} & \multirow{7}{*}{$-$} & 0.508 & 0.586 & 0.733 & 0.800 \\
& $0.2$ & & & 0.514 & 0.596 & 0.737 & 0.808 \\
& $0.4$ & & & 0.521 & 0.605 & 0.738 & 0.807 \\
& $0.6$ & & & 0.528 & 0.615 & 0.739 & 0.810 \\
& $0.8$ & & & \bf 0.532 & \bf 0.619 & \bf 0.743 & \bf 0.814 \\
& $0.9$ & & & 0.525 &  0.612 &  0.740 &  0.811 \\
& $1.0$ & & & 0.391 & 0.542 & 0.576 & 0.749 \\
\hline
\multirow{2}{*}{G2} 
& \multirow{2}{*}{$0.8$} &\textit{Att-Cmp} & \multirow{2}{*}{$-$} & 0.528 & 0.610 & 0.741 & 0.812 \\
& & \textit{Cmp-Att} & & \bf 0.534 & \bf 0.621 & \bf 0.745 & \bf 0.817 \\
\hline
\multirow{3}{*}{G3}
& \multirow{3}{*}{$0.8$} & \multirow{3}{*}{\textit{Cmp-Att}} & $1$ & 0.526 & 0.614 & 0.744 & 0.815 \\
& & & $5$ & \bf 0.537 & \bf 0.624 & \bf 0.746 & \bf 0.817 \\
& & & $10$ & 0.533 & 0.620 & 0.745 & 0.817 \\
\hline
G4 & $0.0$ & \textit{Cmp}  & $5$ & 0.513 & 0.592 & 0.738 & 0.805 \\
\end{tabu}
\end{center}}
\caption{Ablation study of different modules and setups. $\lambda$ is fusion weight of classification scores from 2 subnets. \textit{Cmp}, \textit{Att} represent \textit{compresson module} and \textit{attention module} respectively. $T$ denotes the update interval for short-term, and $-$ means no update is applied. } 
\label{ablation}
\end{table} 

\begin{figure}[!t]
\centering
\subfigure[Search]{
\includegraphics[width=1.6cm]{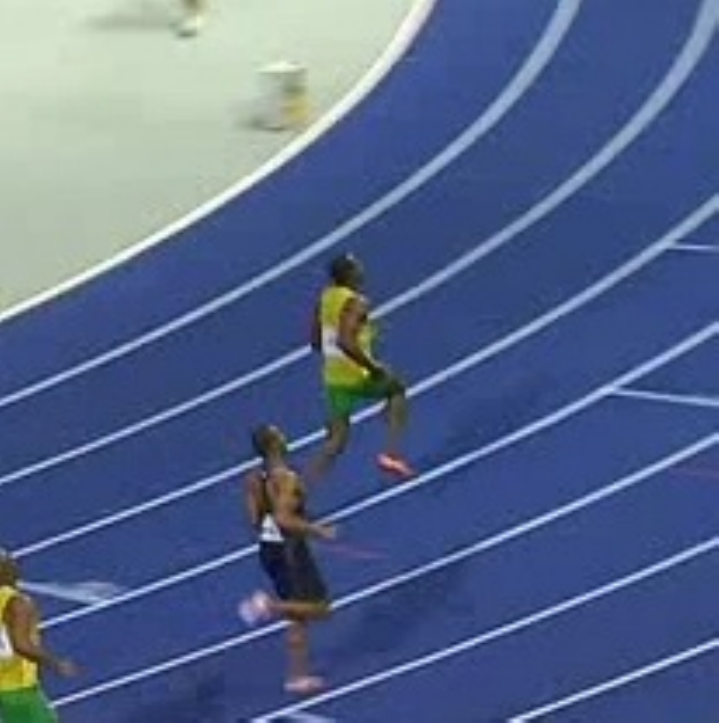}
\label{a}
}
\quad
\subfigure[w/o C]{
\includegraphics[width=1.6cm]{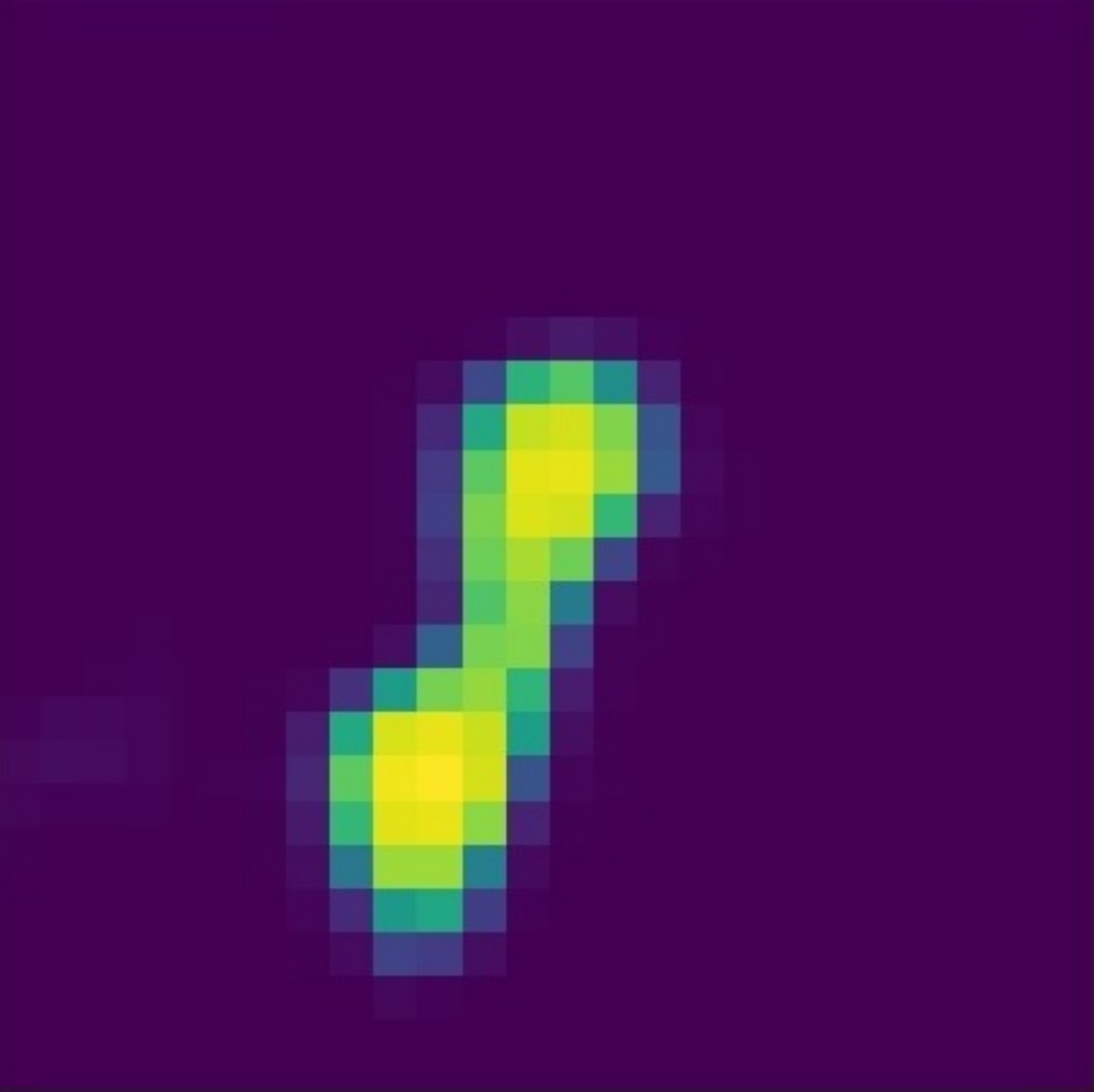}
\label{b}
}
\quad
\subfigure[w/o Att]{
\includegraphics[width=1.6cm]{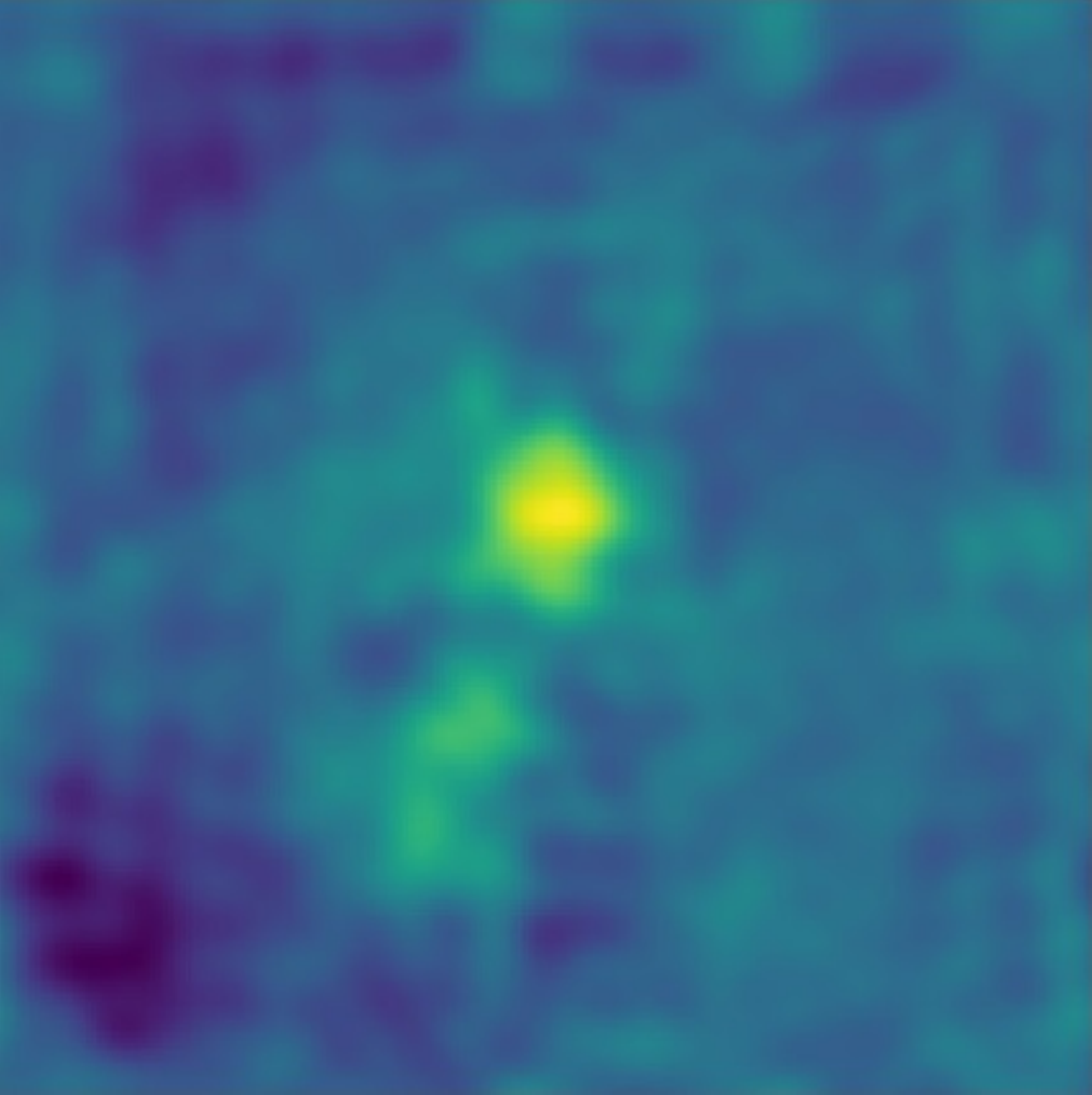}
\label{c}
}
\quad
\subfigure[w/ Att]{
\includegraphics[width=1.6cm]{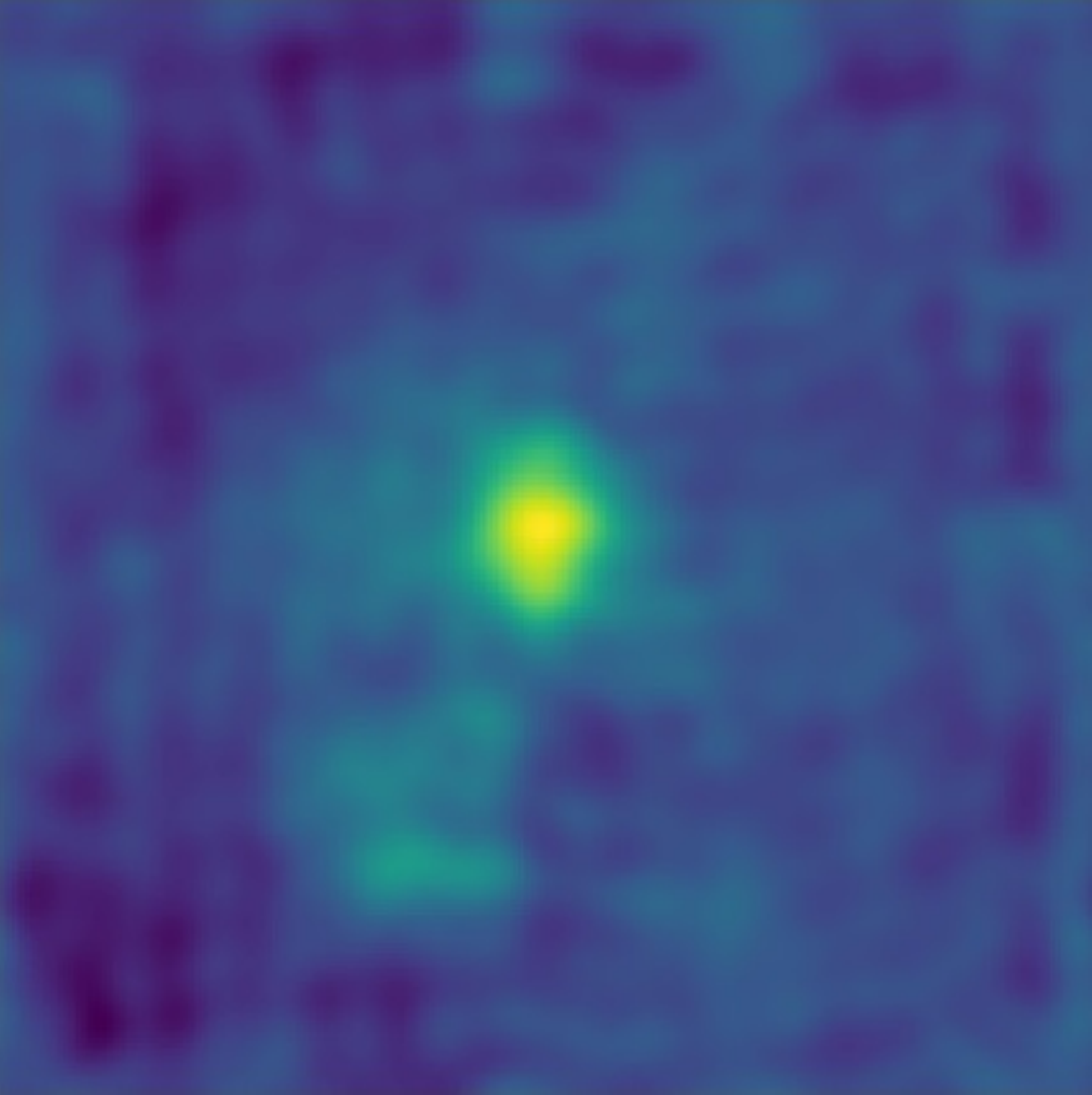}
\label{d}
}
\caption{Visualization on final classification output after the Classification subnet \textit{C} and attention module is used on sequence \textit{Bolt} of OTB2015.}
\end{figure}

\begin{figure}[!t]
\centering
\subfigure{
\includegraphics[width=3.8cm]{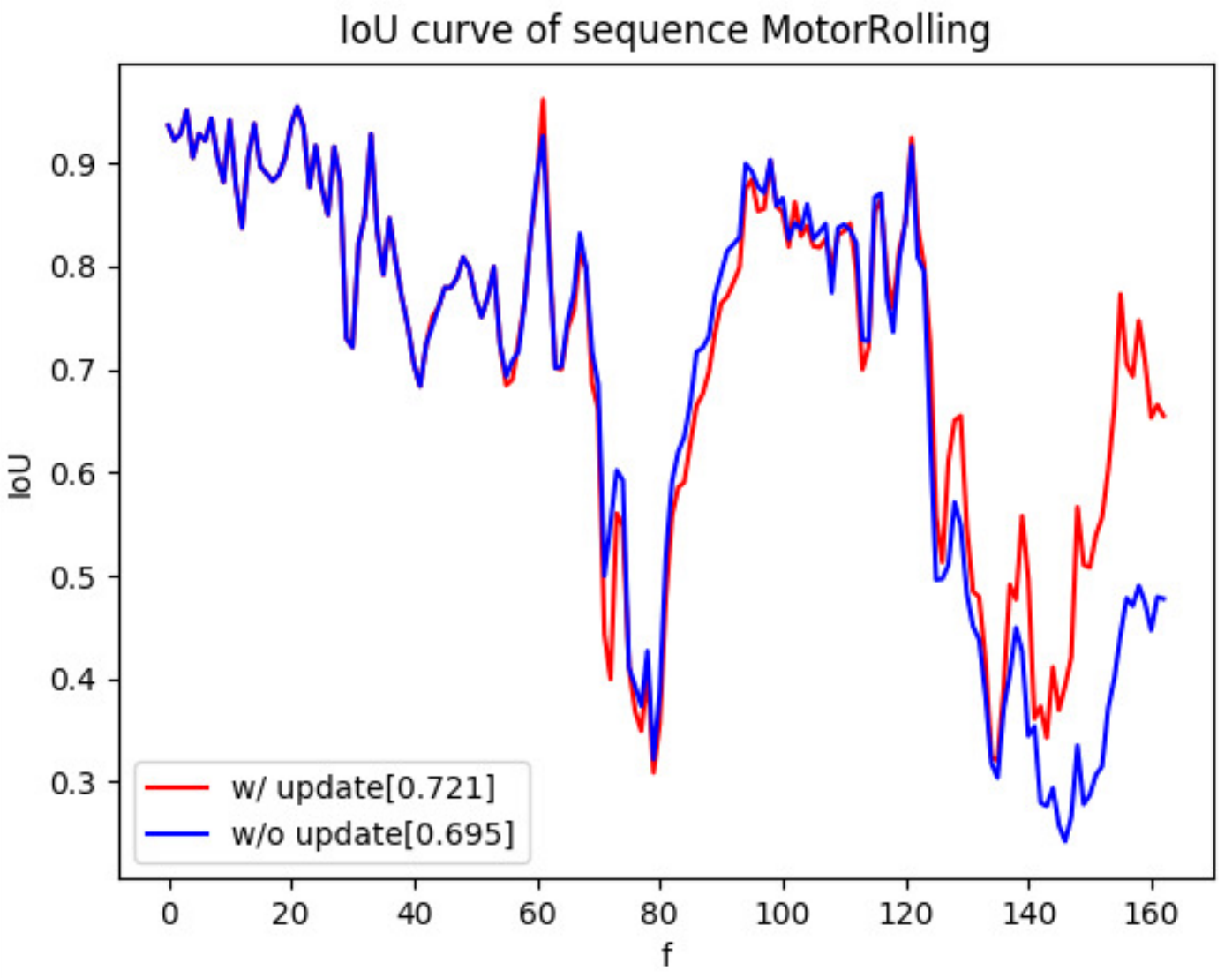}
\label{aa}
}
\quad
\subfigure{
\includegraphics[width=3.8cm]{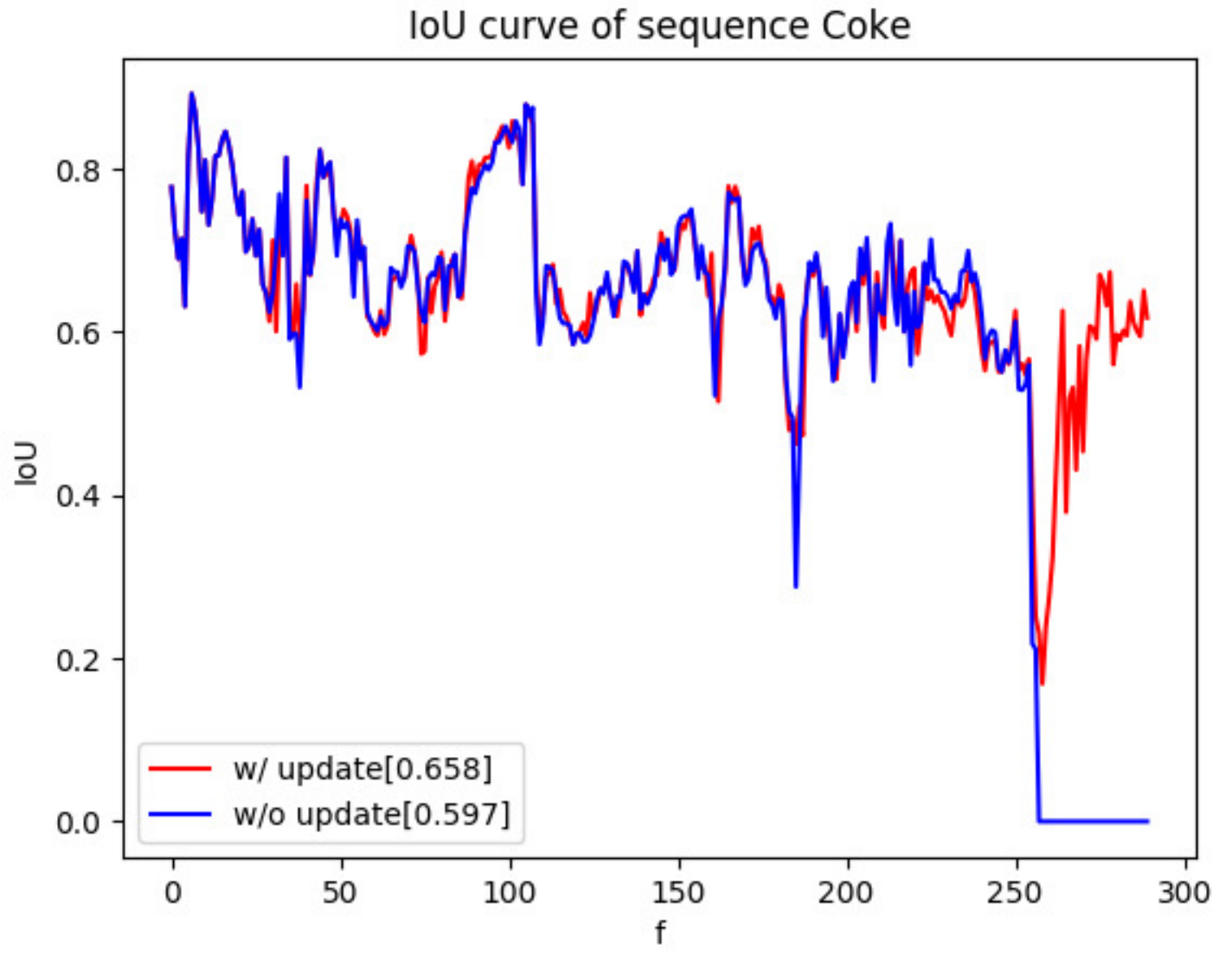}
\label{bb}
}
\caption{IoU curve and expected average IoU on sequence \textit{MotorRolling} (left) and \textit{Coke} (right) of OTB2015.}
\end{figure}

\noindent\textbf{Classification score fusion (G1).} As data reveals, there's a synchronous need of scores from two subnets as two extremes with $\lambda$ being $0$ or $1$ induce poor results while the optimal coefficient falls around $0.8$. Theoretically, it can be partly explained by that, though discriminative enough, online classification score only locates a rough location, while more fine-grained details distinguishing the quality of dense bounding boxes around the peak region need to be fleshed out by the siamese classification score.

\noindent\textbf{Target-specific feature learning (G2).} 
% To unleash the discriminative capability of classification to the largest extent, we fully investigate the impact of each module in our online classification subnet. The compression module cuts down the dimension of the feature map from ResNet50 backbone to cater for fast speed. 
With the best settings in G1, we further delve into the investigation of the proposed attention mechanism. As shown in Table \ref{ablation}, the existence of the attention module further improves trackers' performance. Qualitatively, in the comparison between Figure \ref{b} and Figure \ref{c}, the online classification score is more discriminative to deal with distractors and occlusions. The attention module uplifts this discriminative capability given the fact the peak on ground-truth target is more focused as shown in Figure \ref{c} and Figure \ref{d}. 

\noindent\textbf{Template update interval (G3).} From Table \ref{ablation}, compared with directly updating the short-term sample once it obtains a qualified classification score by setting $T=1$, we find that properly prolonging the update interval and selecting one with the highest classification confidence after sorting allows us a more trust-worthy template. To demonstrate the validity of the update strategy, we provide IoU curve of $2$ sequences. As Figure \ref{aa} shown, short-term template improves the performance especially when putative box overlaps the ground truth not well, and Figure \ref{bb} shows such strategy can even avoid target drift in the long run. 

\noindent\textbf{Validity of update strategy (G4).} To evaluate the effectiveness of our update strategy, G4 is conducted by only using the online classification score to guide template update without fusing it with siamese classification score. Compared with the baseline (see the first column in G1), the validity of updater can be more clearly observed. 

\section{Conclusion}
In this work, we propose an online module with a dual attention mechanism for siamese visual tracking to extract video-specific features under a quadratic problem. The discriminative learning is performed by updating the filter module during tracking, which enhances siamese network's discriminative ability to deal with distractors. And the robust learning is achieved by an extra short-term template branch updating dynamically based on scores from the classifier to handle huge deformation. Without bells and whistles, our proposed online module can easily work with any siamese baselines without degenerating its speed too much. Extensive experiments on 6 major benchmarks are performed to demonstrate the validity and superiority of our approach. 

\bibliographystyle{aaai}
\bibliography{egbib}

\begin{thebibliography}{}

\bibitem[\protect\citeauthoryear{Bertinetto \bgroup et al\mbox.\egroup
  }{2016a}]{SiamFC}
Bertinetto, L.; Valmadre, J.; Henriques, J.~F.; Vedaldi, A.; and Torr, P.~H.
\newblock 2016a.
\newblock Fully-convolutional siamese networks for object tracking.
\newblock In {\em European conference on computer vision},  850--865.
\newblock Springer.

\bibitem[\protect\citeauthoryear{Bertinetto \bgroup et al\mbox.\egroup
  }{2016b}]{CFNet}
Bertinetto, L.; Valmadre, J.; Henriques, J.~F.; Vedaldi, A.; and Torr, P.~H.
\newblock 2016b.
\newblock Fully-convolutional siamese networks for object tracking.
\newblock In {\em European conference on computer vision},  850--865.
\newblock Springer.

\bibitem[\protect\citeauthoryear{Bhat \bgroup et al\mbox.\egroup }{2019}]{DiMP}
Bhat, G.; Danelljan, M.; Van~Gool, L.; and Timofte, R.
\newblock 2019.
\newblock Learning discriminative model prediction for tracking.
\newblock {\em arXiv preprint arXiv:1904.07220}.

\bibitem[\protect\citeauthoryear{Choi, Kwon, and Lee}{2017a}]{UpdateNet}
Choi, J.; Kwon, J.; and Lee, K.~M.
\newblock 2017a.
\newblock Deep meta learning for real-time visual tracking based on
  target-specific feature space.
\newblock {\em arXiv preprint arXiv:1712.09153}.

\bibitem[\protect\citeauthoryear{Choi, Kwon, and Lee}{2017b}]{MLT}
Choi, J.; Kwon, J.; and Lee, K.~M.
\newblock 2017b.
\newblock Deep meta learning for real-time visual tracking based on
  target-specific feature space.
\newblock {\em arXiv preprint arXiv:1712.09153}.

\bibitem[\protect\citeauthoryear{Danelljan \bgroup et al\mbox.\egroup
  }{2017}]{ECO}
Danelljan, M.; Bhat, G.; Shahbaz~Khan, F.; and Felsberg, M.
\newblock 2017.
\newblock Eco: efficient convolution operators for tracking.
\newblock In {\em Proceedings of the IEEE conference on computer vision and
  pattern recognition},  6638--6646.

\bibitem[\protect\citeauthoryear{Danelljan \bgroup et al\mbox.\egroup
  }{2019}]{ATOM}
Danelljan, M.; Bhat, G.; Khan, F.~S.; and Felsberg, M.
\newblock 2019.
\newblock Atom: Accurate tracking by overlap maximization.
\newblock In {\em The IEEE Conference on Computer Vision and Pattern
  Recognition (CVPR)}.

\bibitem[\protect\citeauthoryear{Fan and Ling}{2019}]{C-RPN}
Fan, H., and Ling, H.
\newblock 2019.
\newblock Siamese cascaded region proposal networks for real-time visual
  tracking.
\newblock In {\em The IEEE Conference on Computer Vision and Pattern
  Recognition (CVPR)}.

\bibitem[\protect\citeauthoryear{Fan \bgroup et al\mbox.\egroup }{2019}]{LaSOT}
Fan, H.; Lin, L.; Yang, F.; Chu, P.; Deng, G.; Yu, S.; Bai, H.; Xu, Y.; Liao,
  C.; and Ling, H.
\newblock 2019.
\newblock Lasot: A high-quality benchmark for large-scale single object
  tracking.
\newblock In {\em Proceedings of the IEEE Conference on Computer Vision and
  Pattern Recognition},  5374--5383.

\bibitem[\protect\citeauthoryear{Guo \bgroup et al\mbox.\egroup }{2017}]{DSiam}
Guo, Q.; Feng, W.; Zhou, C.; Huang, R.; Wan, L.; and Wang, S.
\newblock 2017.
\newblock Learning dynamic siamese network for visual object tracking.
\newblock In {\em Proceedings of the IEEE International Conference on Computer
  Vision},  1763--1771.

\bibitem[\protect\citeauthoryear{He \bgroup et al\mbox.\egroup
  }{2018}]{Siam-BM}
He, A.; Luo, C.; Tian, X.; and Zeng, W.
\newblock 2018.
\newblock Towards a better match in siamese network based visual object
  tracker.
\newblock In {\em Proceedings of the European Conference on Computer Vision
  (ECCV)},  0--0.

\bibitem[\protect\citeauthoryear{Henriques \bgroup et al\mbox.\egroup
  }{2014}]{KCF}
Henriques, J.~F.; Caseiro, R.; Martins, P.; and Batista, J.
\newblock 2014.
\newblock High-speed tracking with kernelized correlation filters.
\newblock {\em IEEE transactions on pattern analysis and machine intelligence}
  37(3):583--596.

\bibitem[\protect\citeauthoryear{Kalal, Mikolajczyk, and Matas}{2011}]{TLD}
Kalal, Z.; Mikolajczyk, K.; and Matas, J.
\newblock 2011.
\newblock Tracking-learning-detection.
\newblock {\em IEEE transactions on pattern analysis and machine intelligence}
  34(7):1409--1422.

\bibitem[\protect\citeauthoryear{Kristan \bgroup et al\mbox.\egroup
  }{2018}]{VOT2018}
Kristan, M.; Leonardis, A.; Matas, J.; Felsberg, M.; Pflugfelder, R.;
  Cehovin~Zajc, L.; Vojir, T.; Bhat, G.; Lukezic, A.; Eldesokey, A.; et~al.
\newblock 2018.
\newblock The sixth visual object tracking vot2018 challenge results.
\newblock In {\em Proceedings of the European Conference on Computer Vision
  (ECCV)},  0--0.

\bibitem[\protect\citeauthoryear{Li \bgroup et al\mbox.\egroup
  }{2018}]{SiamRPN}
Li, B.; Wu, W.; Zhu, Z.; Yan, J.; and Hu, X.
\newblock 2018.
\newblock High performance visual tracking with siamese region proposal
  network.
\newblock {\em IEEE Conference on Computer Vision and Pattern Recognition
  (CVPR)}.

\bibitem[\protect\citeauthoryear{Li \bgroup et al\mbox.\egroup
  }{2019a}]{SiamRPN++}
Li, B.; Wu, W.; Wang, Q.; Zhang, F.; Xing, J.; and Yan, J.
\newblock 2019a.
\newblock Siamrpn++: Evolution of siamese visual tracking with very deep
  networks.
\newblock {\em IEEE Conference on Computer Vision and Pattern Recognition
  (CVPR)}.

\bibitem[\protect\citeauthoryear{Li \bgroup et al\mbox.\egroup }{2019b}]{TADT}
Li, X.; Ma, C.; Wu, B.; He, Z.; and Yang, M.-H.
\newblock 2019b.
\newblock Target-aware deep tracking.
\newblock In {\em Proceedings of the IEEE Conference on Computer Vision and
  Pattern Recognition},  1369--1378.

\bibitem[\protect\citeauthoryear{Mueller, Smith, and Ghanem}{2016}]{UAV123}
Mueller, M.; Smith, N.; and Ghanem, B.
\newblock 2016.
\newblock A benchmark and simulator for uav tracking.
\newblock In {\em European conference on computer vision},  445--461.
\newblock Springer.

\bibitem[\protect\citeauthoryear{Muller \bgroup et al\mbox.\egroup
  }{2018}]{TrackingNet}
Muller, M.; Bibi, A.; Giancola, S.; Alsubaihi, S.; and Ghanem, B.
\newblock 2018.
\newblock Trackingnet: A large-scale dataset and benchmark for object tracking
  in the wild.
\newblock In {\em Proceedings of the European Conference on Computer Vision
  (ECCV)},  300--317.

\bibitem[\protect\citeauthoryear{Nam and Han}{2016}]{MDNet}
Nam, H., and Han, B.
\newblock 2016.
\newblock Learning multi-domain convolutional neural networks for visual
  tracking.
\newblock In {\em Proceedings of the IEEE conference on computer vision and
  pattern recognition},  4293--4302.

\bibitem[\protect\citeauthoryear{Park and Berg}{2018}]{Meta-tracker}
Park, E., and Berg, A.~C.
\newblock 2018.
\newblock Meta-tracker: Fast and robust online adaptation for visual object
  trackers.
\newblock {\em Lecture Notes in Computer Science}  587–604.

\bibitem[\protect\citeauthoryear{Wang \bgroup et al\mbox.\egroup
  }{2018}]{RASNet}
Wang, Q.; Teng, Z.; Xing, J.; Gao, J.; Hu, W.; and Maybank, S.
\newblock 2018.
\newblock Learning attentions: residual attentional siamese network for high
  performance online visual tracking.
\newblock In {\em Proceedings of the IEEE Conference on Computer Vision and
  Pattern Recognition},  4854--4863.

\bibitem[\protect\citeauthoryear{Wang \bgroup et al\mbox.\egroup
  }{2019}]{SiamMask}
Wang, Q.; Zhang, L.; Bertinetto, L.; Hu, W.; and Torr, P.~H.
\newblock 2019.
\newblock Fast online object tracking and segmentation: A unifying approach.
\newblock {\em IEEE Conference on Computer Vision and Pattern Recognition
  (CVPR)}.

\bibitem[\protect\citeauthoryear{Wu, Lim, and Yang}{2015}]{OTB2015}
Wu, Y.; Lim, J.; and Yang, M.-H.
\newblock 2015.
\newblock Object tracking benchmark.
\newblock {\em IEEE Transactions on Pattern Analysis and Machine Intelligence}
  37(9):1834--1848.

\bibitem[\protect\citeauthoryear{Yang and Chan}{2018}]{MemTrack}
Yang, T., and Chan, A.~B.
\newblock 2018.
\newblock Learning dynamic memory networks for object tracking.
\newblock In {\em Proceedings of the European Conference on Computer Vision
  (ECCV)},  152--167.

\bibitem[\protect\citeauthoryear{Yang \bgroup et al\mbox.\egroup
  }{2019}]{HASiam}
Yang, K.; Song, H.; Zhang, K.; and Liu, Q.
\newblock 2019.
\newblock Hierarchical attentive siamese network for real-time visual tracking.
\newblock {\em Neural Computing and Applications}  1--12.

\bibitem[\protect\citeauthoryear{Zhang and Peng}{2019}]{SiamDW}
Zhang, Z., and Peng, H.
\newblock 2019.
\newblock Deeper and wider siamese networks for real-time visual tracking.
\newblock In {\em Proceedings of the IEEE Conference on Computer Vision and
  Pattern Recognition},  4591--4600.

\bibitem[\protect\citeauthoryear{Zhang \bgroup et al\mbox.\egroup
  }{2018}]{MBMD}
Zhang, Y.; Wang, D.; Wang, L.; Qi, J.; and Lu, H.
\newblock 2018.
\newblock Learning regression and verification networks for long-term visual
  tracking.
\newblock {\em arXiv preprint arXiv:1809.04320}.

\bibitem[\protect\citeauthoryear{Zhu \bgroup et al\mbox.\egroup
  }{2018}]{DaSiamRPN}
Zhu, Z.; Wang, Q.; Li, B.; Wei, W.; Yan, J.; and Hu, W.
\newblock 2018.
\newblock Distractor-aware siamese networks for visual object tracking.
\newblock {\em The European Conference on Computer Vision (ECCV)}.

\end{thebibliography}

\end{document}